\documentclass[10pt,twocolumn,letterpaper]{article}

\usepackage{cvpr}              % To produce the CAMERA-READY version
\newcommand{\tocite}[1][]{%
  \textcolor{red}{[cite%
  \if\relax\detokenize{#1}\relax\else: #1\fi]}%
}

\newcommand{\sysname}[0]{Scal3R}

\usepackage[normalem]{ulem}

\usepackage{microtype}

\definecolor{cvprblue}{rgb}{0.21,0.49,0.74}
\usepackage[english]{babel}
\usepackage{colortbl}
\usepackage{xcolor} 
\usepackage{tabularx}
\usepackage{multirow}
\usepackage{booktabs}
\usepackage{graphicx}
\usepackage{array}
\usepackage{caption}
\usepackage{amsmath}
\usepackage{float}
\usepackage{booktabs}
\usepackage{capt-of}
\usepackage{ulem}
\usepackage[pagebackref,breaklinks,colorlinks,allcolors=cvprblue]{hyperref}
\usepackage[accsupp]{axessibility}  % Improves PDF readability for those with disabilities.

\def\paperID{***} % *** Enter the Paper ID here
\def\confName{CVPR}
\def\confYear{2026}

\title{Scal3R: Scalable Test-Time Training for Large-Scale 3D Reconstruction}
\author{
    Tao Xie$^{1,2}$ \quad
    Peishan Yang$^{1}$ \quad
    Yudong Jin$^{1}$ \quad
    Yingfeng Cai$^{2}$ \quad
    Wei Yin$^{2}$ \quad
    Weiqiang Ren$^{2}$ \quad \\
    Qian Zhang$^{2}$ \quad
    Wei Hua$^{3}$ \quad
    Sida Peng$^{1}$ \quad
    Xiaoyang Guo$^{2 \dagger}$ \quad
    Xiaowei Zhou$^{1 \dagger}$
    \vspace{0.2cm}
\\
    {\normalsize $^{1}$ Zhejiang University} \quad
    {\normalsize $^{2}$ Horizon Robotics} \quad
    {\normalsize $^{3}$ Zhejiang Lab}
}

\begin{document}
% \maketitle

\twocolumn[{
    \renewcommand\twocolumn[1][]{#1}
    \vspace{-3em}
    \maketitle
    \begin{center}
        \captionsetup{type=figure}
        \centering
        \includegraphics[width=1.0\textwidth]{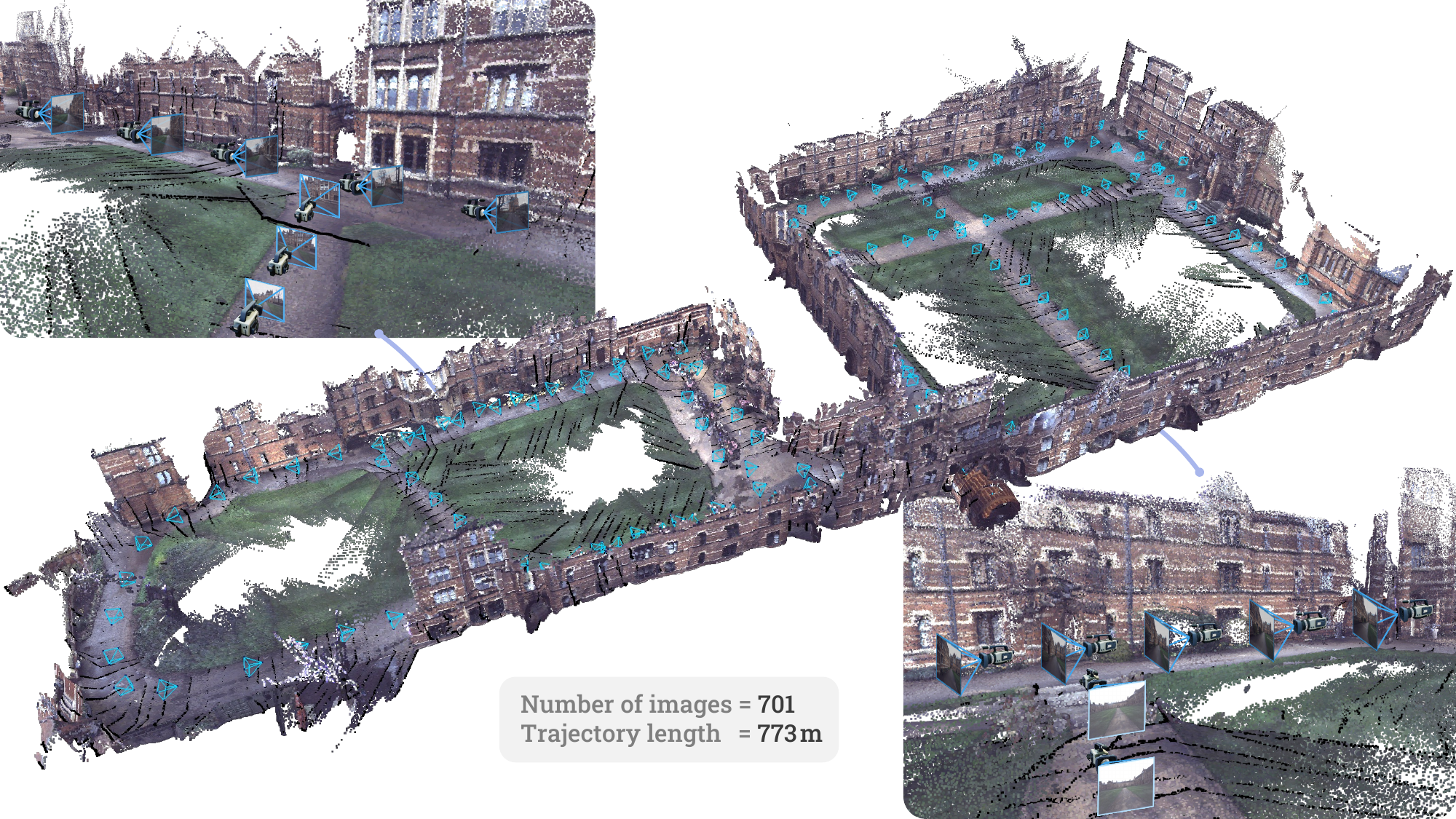}
        \vspace{-0.2in}
        \caption{Large-scale reconstruction on Oxford Spires dataset~\cite{tao2025spires}. \sysname{} reconstructs large-scale 3D scenes from long RGB sequences within a unified inference pipeline, yielding high reconstruction accuracy and efficiency on kilometer-scale scenes.}
        \label{fig:teaser}
    \end{center}
}]

\newcommand\blfootnote[1]{%
  \begingroup
  \renewcommand\thefootnote{}\footnote{#1}%
  \addtocounter{footnote}{-1}%
  \endgroup
}
\blfootnote{\textcolor{black}*The authors from Zhejiang University are affiliated with the State Key Lab of CAD\&CG. {$\dagger$} Co-corresponding authors: Xiaoyang Guo, Xiaowei Zhou}

\begin{abstract}

This paper addresses the task of large-scale 3D scene reconstruction from long video sequences.
Recent feed-forward reconstruction models have shown promising results by directly regressing 3D geometry from RGB images without explicit 3D priors or geometric constraints. However, these methods often struggle to maintain reconstruction accuracy and consistency over long sequences due to limited memory capacity and the inability to effectively capture global contextual cues.
In contrast, humans can naturally exploit the global understanding of the scene to inform local perception.
Motivated by this, we propose a novel neural global context representation that efficiently compresses and retains long-range scene information, enabling the model to leverage extensive contextual cues for enhanced reconstruction accuracy and consistency.
The context representation is realized through a set of lightweight neural sub-networks that are rapidly adapted during test time via self-supervised objectives, which substantially increases memory capacity without incurring significant computational overhead.
The experiments on multiple large-scale benchmarks, including the KITTI Odometry~\cite{Geiger2012CVPR} and Oxford Spires~\cite{tao2025spires} datasets, demonstrate the effectiveness of our approach in handling ultra-large scenes, achieving leading pose accuracy and state-of-the-art 3D reconstruction accuracy while maintaining efficiency.
Code is available at https://zju3dv.github.io/scal3r.

\end{abstract}
    
\section{Introduction}
\label{sec:introduction}

Large-scale 3D scene reconstruction plays a critical role in autonomous driving, robotics mapping, and digital twin modeling. 
Unlike object-centric or small-scale scenes, reconstructing entire environments that span kilometers brings distinct challenges, such as aligning thousands of viewpoints, integrating vastly varying depth and lighting conditions, and preserving both global consistency and fine local details.
While traditional methods aim at large-scale reconstruction, they generally assume known camera intrinsic~\cite{mur2015orb,gao2018ldso,teed2021droid}, or rely on auxiliary sensors (\eg, IMU~\cite{qin2018vins,zhang2015visual,qin2018relocalization}, LiDAR~\cite{zhang2014loam,cwian2021large}) and complex multi-stage workflows~\cite{deng2025gigaslam}, which restricts their flexibility.

By contrast, the field has recently witnessed substantial advances in feed-forward models~\cite{wang2024dust3r,leroy2024grounding,yang2025fast3r,cabon2025must3r,wang2025continuous,wang2025vggt} that directly regress scene geometry from multi-view RGB images.
Among these, VGGT~\cite{wang2025vggt} stands out by adopting a unified Transformer ~\cite{vaswani2017attention} architecture to estimate camera parameters, depth maps, and point clouds in a single pass, yielding high reconstruction accuracy with low computation cost and scaling capability.
However, the attention mechanism’s quadratic computation cost constrains their scalability to ultra-long large-scale sequences.

FastVGGT~\cite{shen2025fastvggt} addresses the computational cost with a token-merging technique~\cite{bolya2023token}, which reduces attention redundancy and enables processing of larger image collections.
However, aggressive token compression discards fine-grained spatial cues and weakens long-range dependencies, thereby undermining global structural consistency and performance.
VGGT-Long~\cite{deng2025vggt}, an alternative approach to improve VGGT~\cite{wang2025vggt}, instead adopts a divide-and-conquer strategy that divides the whole sequence into overlapping chunks, reconstructs each with VGGT, and aligns adjacent chunks into a unified reconstruction.
While this strategy alleviates the quadratic computation overhead, its alignment is highly sensitive to local accuracy.
As each chunk is processed independently without global context, local prediction errors in large, complex scenes or with limited observations often lead to degraded performance.

These observations highlight a key question: \textit{how can a 3D foundation model, akin to human perception, efficiently retain and leverage long-term contextual cues to improve the reconstruction accuracy across large-scale scenes?}
To address this challenge, our key idea is to develop a global context representation that effectively compresses and stores long-term scene context, coupled with an efficient aggregation and sharing mechanism to exploit this context during reconstruction.
By bridging local observations with global context, this design improves local accuracy and enables scalable large-scale 3D reconstruction.

To this end, we present \sysname{}, a novel framework for reconstructing high-quality kilometer-scale 3D scenes from RGB-only sequences.
Building upon VGGT~\cite{wang2025vggt}'s strong visual geometry reasoning capability, we address the loss of global information inherent in chunk-wise processing~\cite{deng2025vggt} by introducing a neural global context representation for sequence-level information aggregation.
Inspired by recent advances in subquadratic sequence modeling~\cite{wang2025test,zhang2025test}, we realize this representation with a set of online-adapted, lightweight sub-networks that efficiently aggregate long-range context during inference via self-supervised objectives.
The resulting neural global context representation offers strong expressive capacity to compactly encode and preserve extensive context, effectively mitigating the long-range dependencies degradation caused by feature over-compression.

However, a global context store alone is not enough, it must be exploited to enhance reconstruction.
We therefore design a context aggregation mechanism built on our neural global context that, at test time, coordinates the self-supervised online adaptation of the lightweight sub-networks so that global cues are aggregated and shared across the entire sequence.
Together, the representation and the aggregation endow local reconstruction with richer global priors, reducing sensitivity to sparse or ambiguous views.
This coupling yields substantially better local accuracy and consistency while preserving the scalability and efficiency of VGGT~\cite{wang2025vggt}, enabling large-scale training on diverse datasets.

Extensive experiments on Virtual KITTI~\cite{cabon2020virtual}, and zero-shot evaluations on KITTI~\cite{Geiger2012CVPR} and Oxford Spires~\cite{tao2025spires} demonstrate \sysname{}'s state-of-the-art pose estimation accuracy, showcasing its effectiveness in ultra-long sequence handling.
Additional 3D reconstruction evaluations corroborate \sysname{}’s robustness and geometric accuracy across diverse scenes, as illustrated in Section~\ref{sec:experiment}.

In summary, we make the following contributions.
\begin{itemize}
    \item We present \sysname{}, a novel framework capable of reconstructing high-quality kilometer-scale 3D scenes from RGB-only sequences.
    \item We introduce a global context representation together with a context aggregation mechanism that jointly compresses, retains, and shares long-term information across sequences, enabling globally consistent and scalable reconstruction over vast environments.
    \item Extensive evaluations on diverse large-scale datasets demonstrate that \sysname{} achieves state-of-the-art performance with superior accuracy and global consistency.
\end{itemize}

\section{Related Work}
\label{sec:related_work}

\noindent\textbf{SfM and SLAM. }
Classical structure-from-motion (SfM) methods ~\cite{snavely2006photo,agarwal2011building,schonberger2016structure,wilson2014robust,cui2015global,pan2024global} estimate camera poses and 3D structure through feature matching, triangulation, and bundle adjustment (BA). While accurate, they often fail in textureless areas or repetitive patterns where reliable feature correspondences are scarce.
Learning-based extensions enhance robustness and scalability by integrating neural feature detection ~\cite{detone2018superpoint,dusmanu2019d2,tyszkiewicz2020disk,yi2016lift}, matching ~\cite{chen2021learning,lindenberger2023lightglue,sun2021loftr,wang2024eloftr,he2024dfsfm}, or more robust 3D geometric representations~\cite{xu2024towards}, yet they still depend on expensive global optimization and struggle to scale to long trajectories or complex scenes.
More recent end-to-end approaches proposed to directly regress~\cite{parameshwara2022diffposenet,zhang2022relpose,deng2025sail} or solve poses via differentiable BA~\cite{tang2018ba,gu2021dro,wang2024vggsfm} or optimization~\cite{lin2025longsplat}, while eliminating explicit feature matching, they have scalability or efficiency limitations on real-world settings.
Visual simultaneous localization and mapping (SLAM) methods, in contrast, estimate camera poses and build maps incrementally, achieving real-time performance~\cite{mur2015orb,gao2018ldso,teed2021droid,teed2023deep,lipson2024deep,peng2025gaussian}. However, they typically rely on known camera intrinsics ~\cite{mur2015orb,gao2018ldso,teed2021droid,teed2023deep,lipson2024deep} or auxiliary sensors ~\cite{qin2018vins,zhang2015visual,qin2018relocalization,zhang2014loam,cwian2021large,jiao2021greedy}, and can be brittle in challenging reflective scenes~\cite{herbert2024benchmarking}, which limits their flexibility in unconstrained settings.

\noindent\textbf{Feed-forward reconstruction models. }
A recent trend is to directly regress 3D geometry from RGB images using feed-forward neural networks, without relying on explicit 3D priors or geometric constraints.
DUSt3R~\cite{wang2024dust3r} and MASt3R~\cite{murai2025mast3r} take early steps in this direction by directly predicting dense pointmaps from uncalibrated image pairs with a Transformer-based architecture, removing the need for known camera intrinsics or poses. However, their two-view design limits scalability to larger scenes.
Subsequent studies~\cite{tang2024mv,cabon2025must3r,yang2025fast3r,zhang2025flarefeedforwardgeometryappearance,wang2025vggt} extend these ideas to multi-view settings.
Among them, VGGT~\cite{wang2025vggt} achieves state-of-the-art performance, but its quadratic attention limits scalability to very long sequences.
Online variants introduce memory state tokens~\cite{wang20243d,wang2025continuous,chen2025ttt3r} or causal Transformer structures~\cite{streamVGGT,lan2025stream3r} to amortize computation over time, but fixed memory size or limited causal horizon still causes drift and accumulated error on long sequences. TTT3R~\cite{chen2025ttt3r} further casts memory update as test-time learning, but still relies on a fixed-size token set.
We instead propose a scalable global context representation with larger memory capacity for long-range dependencies.

\noindent\textbf{Memory mechanisms. }
Modern recurrent neural networks (RNNs), particularly linear-attention~\cite{katharopoulos2020transformers,schmidhuber1992learning} variants such as Mamba~\cite{dao2024transformers,gu2024mamba}, RWKV~\cite{peng2023rwkv}, and DeltaNet~\cite{schlag2021linear,yang2024parallelizing}, provide an efficient alternative to standard quadratic complexity attention for context modeling and have demonstrated impressive performance in natural language tasks.
However, these models compress the entire history into a finite-size hidden state, which limits their ability to capture complex long-range dependencies, especially in tasks like large-scale 3D perception~\cite{wang2025continuous,chen2025ttt3r} and long video generation~\cite{dalal2025one}.
To overcome this limitation, test-time training (TTT) ~\cite{sun2024learning} and its follow-ups~\cite {behrouz2024titans,zhang2025test} have emerged as a promising technique that extends the recurrent state to an online-adapted non-linear network, substantially increasing memory capacity and improving long-term context modelling.
In parallel, other approaches~\cite{wu2025point3r,yu2025context} employ explicit caches or memory banks to store historical features. While these methods mitigate forgetting, they often face practical challenges in controlling memory growth and computational overhead.

\section{Preliminary}
\label{sec:preliminary}

We begin by introducing the preliminary concepts, including VGGT~\cite{wang2025vggt} and Test-Time Training (TTT)~\cite{sun2024learning}.

\subsection{VGGT}
\label{subsec:vggt}

Given an input RGB sequence $\mathcal{I}=\{I_i\in\mathbb{R}^{3\times H\times W}\mid i=1,\ldots,N\}$ observing the same 3D scene, VGGT adopts a unified transformer $f$ to map each frame in the sequence to its corresponding 3D annotations:
\begin{equation}
    f\bigl( \{I_i\}_{i=1}^N \bigr)
    = \{\boldsymbol{c}_i, D_i, P_i, T_i\}_{i=1}^N,
\end{equation}
where $\boldsymbol{c}_i\in\mathbb{R}^{9}$, $D_i\in\mathbb{R}^{1\times H\times W}$, $P_i\in\mathbb{R}^{3\times H\times W}$, and $T_i\in\mathbb{R}^{C\times H\times W}$ denote the camera parameters (intrinsic and extrinsic), depth map, point cloud, and the feature grid for point tracking of frame $I_i$, respectively.

VGGT consists of three core components. First, a DINOv2~\cite{caron2021emerging} encoder that patchfies and extracts features for each frame, which are then concatenated into image tokens $\mathcal{F} = \bigcup_{i=1}^{N} \{ F_i | F_i \in \mathbb{R}^{K \times C} \}$.
These tokens are then processed by a stack of 24 attention layers that alternate between frame-wise self-attention (within each image) and global self-attention (across images). This alternation enables effective modeling of both intra-frame detail and inter-frame geometry consistency.
Finally, multiple dedicated output heads predict the camera parameters, depth maps, point clouds, and feature grids from the processed tokens.

\subsection{Test-Time Training}
\label{subsec:ttt}

Consider a one-dimensional sequence $\{x_t\mid t=1,\ldots,N\}$ of $N$ tokens, where each token $x_t\in\mathbb{R}^d$ is a $d$-dimensional vector.
To alleviate the quadratic complexity of softmax attention, recurrent neural networks (RNNs) and their variants~\cite{hochreiter1997long,gu2024mamba,peng2023rwkv,peng2023rwkv} compress sequence context into a fixed-size hidden state $h_t\in\mathbb{R}^d$.
At each timestamp $t$, the hidden state is updated based on the current input $x_t$ and the previous hidden state $h_{t-1}$ as:
\begin{equation}
    h_t = \sigma(\theta_{ss}h_{t-1} + \theta_{sx}x_{t}).
\end{equation}
However, this compression is inherently constrained by the representational capacity of the hidden state, leading to information degradation, especially in long sequences.

\begin{figure*}[ht!]
    \centering
    \includegraphics[width=1.0\linewidth]{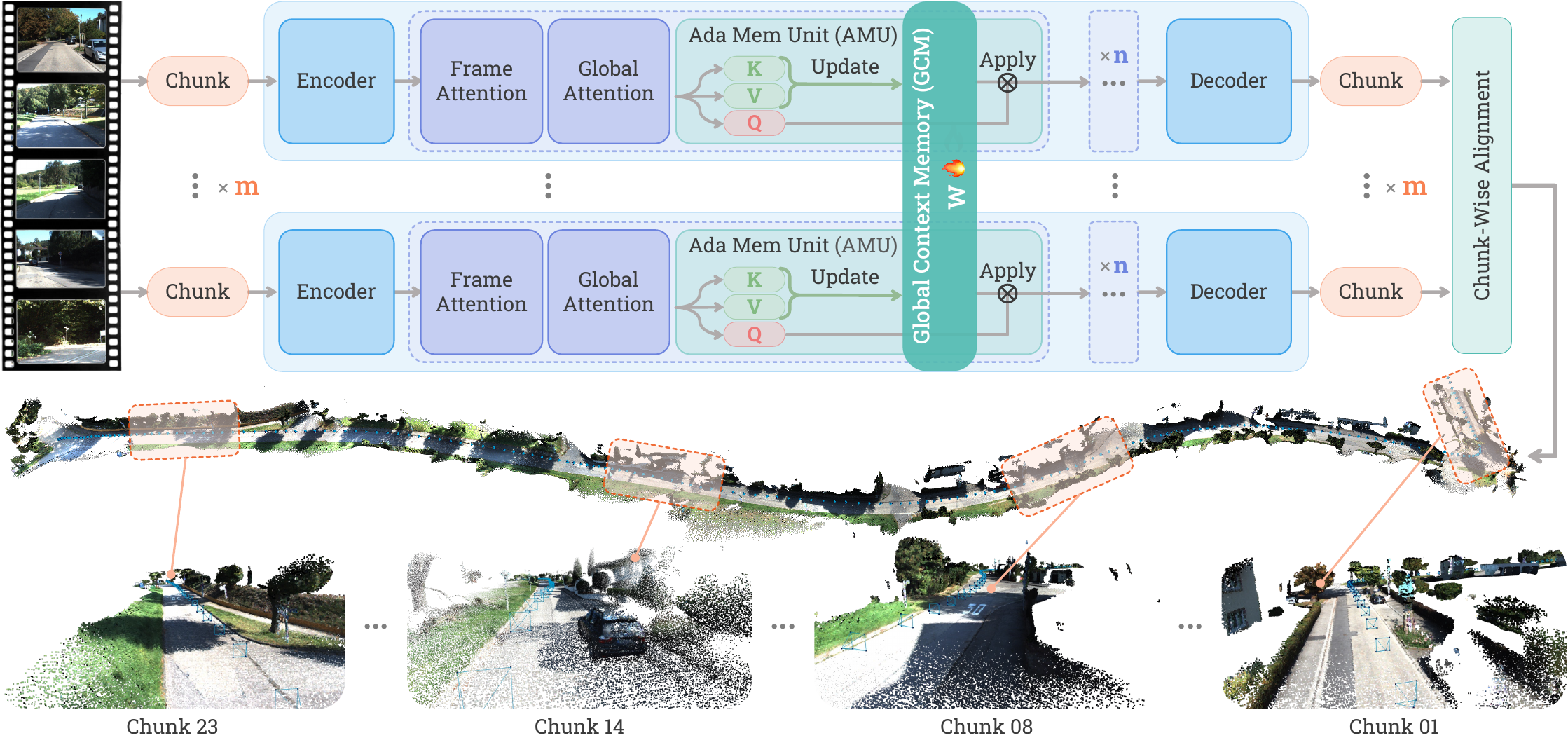}
    \vspace{-15pt}
    \caption{\textbf{Overview of \sysname{}}. Our model takes a long sequence of RGB images as input and reconstructs the 3D scene within a unified inference pipeline. Specifically, the input sequence is divided into overlapping chunks that are processed in parallel across multiple GPUs. Each chunk is processed by our \sysname{} backbone, which incorporates our proposed neural global context representation and aggregation mechanism to capture and share global context across the entire sequence. The resulting camera poses and depth maps from all chunks are then aligned and fused to generate the final 3D reconstruction of the scene.}
    \label{fig:pipeline}
    \vspace{-10pt}
\end{figure*}

To overcome this limitation, Test-Time Training (TTT)~\cite{sun2024learning} introduces \textit{fast weights}, a set of rapidly adaptable neural parameters $W$ that dynamically store contextual information during inference through self-supervised updates~\cite{raina2007self}.
Unlike conventional static model parameters, TTT optimizes \textit{fast weights} in an inner loop to capture contextual dependencies, while the main network parameters are trained in an outer loop for stable generalization.
Following standard attention formulations, each input token $x_t$ is projected into 
query $q$, key $k$, and value $v$, which are learned in the outer loop defining the attention behavior.
While the \textit{fast weights} $W$, updated in the inner loop, serve as dynamic memory that accumulates contextual information over time.
TTT then defines two key operations:
\begin{equation}
    \text{\textbf{update}}: W\leftarrow W-\eta\nabla_{W}\mathcal{L}\left(f_{W}(k),v\right),
\end{equation}
where $\mathcal{L}(\cdot,\cdot)$ is loss function between the transformed key $f_{W}(k)$ and value $v$, encouraging the fast weights to store accurate key–value mappings~\cite{wang2025test}. The updated fast weights $W$ are then used to compute the output $o$ for the current token as:
\begin{equation}
    \text{\textbf{apply}}: o = f_{W}(q).
\end{equation}
By treating the context as an unlabeled dataset and the hidden state as the weights of a machine learning model, TTT effectively enlarges the context capacity beyond fixed-size vectors while retaining the scalability, as shown by recent studies~\cite{zhang2025test}.

\section{Method}
\label{sec:method}

We introduce \sysname{}, a novel framework for kilometer-scale 3D reconstruction from RGB-only sequences.
To address the challenge of thousands of input images in VGGT~\cite{wang2025vggt}, we embed an efficient context aggregation mechanism based on Test-Time Training to capture global contextual cues across entire sequences, while preserving VGGT's strong geometric reasoning capabilities.
Figure~\ref{fig:pipeline} gives an overview of our approach.
In this section, we begin with the overall model architecture (Section~\ref{subsec:architecture}), followed by details of our global context representation and context aggregation mechanism (Section~\ref{subsec:ttt_as_memory}), and finally describe the training and inference procedures (Section~\ref{subsec:training}, \ref{subsec:inference}).

\subsection{Model Overview}
\label{subsec:architecture}

Given a large set of input RGB images $\mathcal{I}$ as defined in Section~\ref{subsec:vggt}, directly applying VGGT~\cite{wang2025vggt} is infeasible due to the quadratic complexity of the attention~\cite{vaswani2017attention} operation.
VGGT-Long~\cite{deng2025vggt} mitigates this issue by partitioning the input sequence into overlapping chunks, processing each chunk independently, and then aligning adjacent results. However, this approach fails to leverage long-range contextual information and is sensitive to local inconsistencies in VGGT's predictions.
Inspired by the recent success of Test-Time Training (TTT)~\cite{sun2024learning,dalal2025one,zhang2025test} in long-context modeling as discussed in Section~\ref{subsec:ttt}, our key insight is to incorporate the TTT modules into VGGT to capture and utilize long-range dependencies across the entire sequence effectively.

To handle the large amount of input images, we first divide the sequence $\mathcal{I}$ into $K$ overlapping chunks $\{\mathcal{I}_k\mid k=1,\ldots,K\}$. Let $M$ be the chunk size and $O$ be the overlap size, then each chunk $\mathcal{I}_k$ contains images $\{I_{(k-1)(M-O)+1}, \ldots, I_{(k-1)(M-O)+M}\}$. These chunks are then distributed across different GPUs and processed by our model in parallel, where the corresponding camera parameters $\mathcal{\boldsymbol{c}}_k$, depth maps $\mathcal{D}_k$, and point clouds $\mathcal{P}_k$ are predicted.

\noindent\textbf{Global Context Memory.}
Following VGGT~\cite{wang2025vggt}, we build our model as a large Transformer\cite{vaswani2017attention} comprising a DINOv2~\cite{caron2021emerging} encoder, alternating attention layers, and multiple output heads for 3D predictions.
At the core of our architecture lies a novel neural Global Context Memory (GCM) module, whose adaptive memory parameters are implemented by several Adaptive Memory Units (AMUs), as illustrated in Figure~\ref{fig:pipeline}.
Each AMU is implemented as a lightweight neural sub-network that is rapidly adapted during inference through self-supervised updates~\cite{sun2024learning}.
The GCM module is attached after the global attention layer to capture and store long-range contextual information, as illustrated in Figure~\ref{fig:pipeline}, we attach 4 GCM modules across our experiments.
Formally, let $\mathcal{X}_{k}^i$ denote the output tokens of the $i$-th global attention layer for chunk $\mathcal{I}_k$.
The GCM module produces the updated tokens as:
\begin{equation}
    \text{gate}(\mathrm{GCM}, \mathcal{X}_k^i; \alpha) = \alpha \otimes \mathrm{GCM}(\mathcal{X}_k^i) + \mathcal{X}_k^i,
\end{equation}
where $\mathrm{GCM}(\cdot)$ denotes the context update and apply operation (detailed in Section~\ref{subsec:ttt_as_memory}), and $\alpha \in \mathbb{R}^d$ is a learnable gate vector that adaptively balances the relative contributions of the GCM output and the original tokens.
We formulate the standard alternating attention operation as:
\begin{equation}
    \bar{\mathcal{X}}_k^i = \text{gattn}\big( \text{fattn}(\mathcal{X}_k^i) \big) + \mathcal{X}_k^i,
\end{equation}
where $\text{fattn}(\cdot)$ and $\text{gattn}(\cdot)$ denote the intra-frame and inter-frame attention operations, respectively.
With the integration of our GCM module, the formulation now becomes:
\begin{equation}
    \bar{\mathcal{X}}_k^i = \text{gate}(\mathrm{GCM}, \text{gattn}\big( \text{fattn}(\mathcal{X}_k^i) \big); \alpha) + \mathcal{X}_k^i,
\end{equation}
and the resulting global-context enhanced tokens are then passed to the dedicated output heads to predict the 3D scene representations for each chunk.

This simple yet effective architecture enables our model to capture long-range dependencies through the GCM modules, while preserving VGGT’s strong geometric reasoning and scalability, making large-scale training across diverse datasets possible.

\begin{table*}[ht!]
    \renewcommand{\arraystretch}{0.8}
    \centering
    \scriptsize
    \setlength{\tabcolsep}{1.2pt}
    \newcolumntype{C}{>{\centering\arraybackslash}m{1.12cm}}

    \caption{\textbf{Camera pose and resource evaluation} on Virtual KITTI~\cite{cabon2020virtual}, KITTI Odometry~\cite{Geiger2012CVPR}, and Oxford Spires~\cite{tao2025spires}. We report RRE ($^\circ$/100m), RTE (m/100m), and ATE (m). Failed scenes are assigned the worst valid score when computing dataset averages. Methods marked with $^\dagger$ require known camera intrinsics. Best and second-best results are shown in \textbf{bold} and \underline{underlined}.}
    \label{tab:pose}

    \resizebox{\textwidth}{!}{%
    \begin{tabular}{@{\hspace{2.8pt}}l*{12}{C}}
            \toprule
            \multirow{3}{*}{\raisebox{-1.8ex}{\textbf{Method}}} & \multicolumn{3}{c}{\textbf{VKITTI2}}                        & \multicolumn{3}{c}{\textbf{KITTI}}                           & \multicolumn{3}{c}{\textbf{Oxford Spires}}                  & \multicolumn{3}{c}{\textbf{Resources}}                      \\[0.35ex]
                                                     & \multicolumn{3}{c}{\scriptsize 223--837 frames, 52--711 meters}    & \multicolumn{3}{c}{\scriptsize 271--4661 frames, 394--5067 meters}  & \multicolumn{3}{c}{\scriptsize 351--787 frames, 280--773 meters}   & \multicolumn{3}{c}{\scriptsize KITTI 03/04/10, avg.\ 758 frames}   \\
            \cmidrule(lr){2-4} \cmidrule(lr){5-7} \cmidrule(lr){8-10} \cmidrule(lr){11-13}
                                                     & RRE $\downarrow$                  & RTE $\downarrow$                  & ATE $\downarrow$                  & RRE $\downarrow$                  & RTE $\downarrow$                  & ATE $\downarrow$                  & RRE $\downarrow$                  & RTE $\downarrow$                  & ATE $\downarrow$                  & Memory $\downarrow$               & Time $\downarrow$                & FPS $\uparrow$                  \\ \specialrule{\lightrulewidth}{0.35ex}{0.85ex}
            MASt3R-SLAM~\cite{murai2025mast3r}       & 15.81                             & 70.48                             & 78.33                             & 22.42                             & 67.72                             & 191.71                            & 59.67                             & 29.82                             & 29.22                             & 6.74                              & 99.30                            & 7.37                            \\
            VGGT-SLAM~\cite{maggio2025vggt}          & 12.92                             & 21.27                             & 17.18                             & 33.27                             & 78.95                             & 214.88                            & 55.60                             & 32.14                             & 26.85                             & 10.67                             & 39.72                            & 19.85                           \\
            StreamVGGT~\cite{streamVGGT}             & 13.47                             & 58.07                             & 68.97                             & 24.06                             & 84.46                             & 226.15                            & 71.28                             & 37.14                             & 34.35                             & 6.66                              & 32.61                            & 23.14                           \\
            STream3R~\cite{lan2025stream3r}          & 13.46                             & 76.06                             & 70.87                             & 24.06                             & 81.63                             & 227.77                            & 71.29                             & 36.65                             & 34.65                             & 4.70                              & 111.23                           & 8.19                            \\
            CUT3R~\cite{wang2025continuous}          & 7.93                              & 40.42                             & 50.75                             & 24.24                             & 73.65                             & 209.78                            & 54.69                             & 32.15                             & 28.01                             & 6.50                              & \underline{22.96}                & \underline{32.87}               \\
            TTT3R~\cite{chen2025ttt3r}               & 5.88                              & 16.34                             & 23.49                             & 21.90                             & 68.55                             & 177.73                            & 62.68                             & 35.51                             & 31.57                             & 4.59                              & 23.65                            & 31.95                           \\
            FastVGGT~\cite{shen2025fastvggt}         & 3.13                              & 38.64                             & 21.83                             & 22.47                             & 69.58                             & 206.69                            & 65.35                             & 37.55                             & 31.18                             & 22.58                             & 48.13                            & 18.22                           \\
            VGGT-Long~\cite{deng2025vggt}            & 0.71                              & 2.01                              & 1.03                              & 1.71                              & \underline{9.67}                  & \underline{25.94}                 & 30.91                             & 20.79                             & 15.46                             & 11.77                             & 168.83                           & 4.80                            \\
            COLMAP~\cite{schonberger2016structure}   & 2.53                              & 7.63                              & 9.09                              & \underline{0.62}                  & 15.88                             & 37.79                             & \textbf{0.32}                     & \textbf{0.24}                     & \textbf{0.15}                     & \textbf{0.00}                     & 6614.73                          & 0.17                            \\
            MASt3R-SfM~\cite{leroy2024grounding}     & 18.28                             & 23.79                             & 40.57                             & 25.43                             & 53.70                             & 171.28                            & 25.83                             & 28.60                             & 25.83                             & 8.04                              & 2766.76                          & 0.27                            \\
            DROID-SLAM\smash{$^\dagger$}~\cite{teed2021droid}& 26.90                     & 41.41                             & 2.47                              & 25.38                             & 58.37                             & 50.71                             & 23.97                             & 45.11                             & 23.97                             & 10.29                             & 56.14                            & 13.58                           \\
            DPVO++\smash{$^\dagger$}~\cite{lipson2024deep}   & \textbf{0.07}             & \textbf{0.39}                     & \textbf{0.48}                     & \textbf{0.23}                     & 15.17                             & 52.69                             & 29.17                             & 30.71                             & 29.17                             & \underline{0.89}                  & \textbf{20.71}                   & \textbf{35.35}                  \\ \specialrule{\lightrulewidth}{0.25ex}{0.55ex}
            Ours                                     & \underline{0.41}                  & \underline{0.78}                  & \underline{0.85}                  & 0.97                              & \textbf{4.61}                     & \textbf{14.55}                    & \underline{7.87}                  & \underline{6.55}                  & \underline{4.45}                  & 10.32                             & 300.76                           & 2.53                            \\ \bottomrule
    \end{tabular}%
    }

    \vspace{-10pt}
\end{table*}

\subsection{Test-Time Training as Memory}
\label{subsec:ttt_as_memory}

Although the adaptable neural sub-networks substantially enlarge memory capacity compared to the fixed-size state token of traditional RNNs (as discussed in Section~\ref{sec:preliminary}), existing TTT~\cite{wang2025test,behrouz2025s,karami2025lattice,sun2024learning} approaches still struggle to scale to long contexts.
This limitation primarily stems from inefficient small-batch updates and sub-optimal GPU utilization, where frequent fine-grained updates hinder throughput and constrain the maximum sequence length.

Inspired by recent work LaCT~\cite{zhang2025test}, which adopts extremely large chunks as the update unit in TTT to improve parallelism and GPU utilization,
we treat all tokens $\mathcal{X}_k$ within each chunk as a single update unit in our Global Context Memory (GCM) module.
This design enables scalable updates of the non-linear Adaptive Memory Units (AMUs) (Section~\ref{subsec:architecture}) within the GCM module, thereby enhancing both memory capacity and computational efficiency during training and inference.
Specifically, as illustrated in Figure~\ref{fig:pipeline}, the GCM module consists of three components: a query-key-value projection layer, a compact MLP network serving as the AMUs, and an output projection layer.
Given the input tokens $\mathcal{X}_k\in\mathbb{R}^{M\times d}$ of chunk $\mathcal{I}_k$, the GCM module first projects them into key and value matrices $K, V \in \mathbb{R}^{M\times d}$, representing the current context.
This context is then encoded into the AMUs $W\in\mathbb{R}^{H}$ through a chunk-wise update operation:
\begin{align}
    W \leftarrow W - \nabla_{W} \sum_{i=1}^{M} \eta_i \mathcal{L}\big(f_{W}(k_i), v_i\big),
\end{align}
where $M$ is the chunk size, and $\eta_i$ is a token-wise learning rate predicted from the input tokens.
We adopt a standard dot-product loss as the self-supervised objective, following the practice of~\cite{sun2024learning,zhang2025test}:
\begin{align}
    \mathcal{L}\big( f_W(K), V \big) = \sum_{i=1}^{M} -f_W(k_i)^\top v_i.
\end{align}
After the update, the AMUs $W$ store the contextual information of the current chunk $\mathcal{I}_k$, which is subsequently used to transform the query tokens $Q\in\mathbb{R}^{M\times d}$ (also projected from $\mathcal{X}_k$) to produce the output tokens $f_{W}(Q)$.

\noindent\textbf{Global Context Synchronization.}
While the GCM module effectively captures intra-chunk context, it remains confined within individual chunks and lacks the ability to exploit sequence-wide global context.
To address this limitation, we introduce a Global Context Synchronization (GCS) mechanism that enables efficient cross-chunk aggregation and exploitation of global context during both training and inference, which is crucial for achieving consistent large-scale 3D reconstruction.
To elaborate, we frame the partitioning of the input image set across different GPUs as a form of context parallelism~\cite{yang2024context}.
Each GPU computes the updates of its local AMUs, after which these updates are synchronized by summing the gradients and broadcasting the result across all GPUs to realize global context sharing.
Formally, the synchronized gradient is expressed as:
\begin{align}
    g = \nabla_{W} \sum_{j=1}^{K} \sum_{i=1}^{M} \eta_i \mathcal{L}_i
    = \sum_{j=1}^{K} \nabla_{W} \sum_{i=1}^{M} \eta_i \mathcal{L}_i
\end{align}
where $K$ is the number of chunks and $M$ is the chunk size.
The aggregated gradient $g$ is then applied to update the adaptive memory unit $W$ on all GPUs.
This operation is efficiently implemented using the \textit{all-reduce} primitives of PyTorch~\cite{paszke2019pytorchimperativestylehighperformance}, ensuring minimal communication overhead during both training and inference.
By doing so, each local chunk is enriched with substantial global observations, which improves local accuracy, strengthens cross-chunk consistency, and elevates overall reconstruction performance.

\subsection{Training}
\label{subsec:training}

\noindent\textbf{Training datasets. }
Our model is trained on the following datasets: Co3Dv2~\cite{reizenstein2021common}, BlendedMVS~\cite{yao2020blendedmvs}, DL3DV~\cite{ling2024dl3dv}, MegaDepth~\cite{li2018megadepth}, WildRGB~\cite{xia2024rgbd}, ScanNet++~\cite{yeshwanth2023scannet++}, HyperSim~\cite{roberts2021hypersim}, Mapillary~\cite{antequera2020mapillary}, Replica~\cite{straub2019replica}, MVS-Synth~\cite{huang2018deepmvs}, Virtual KITTI~\cite{cabon2020virtual}, Aria Synthetic Environments~\cite{pan2023aria}, Aria Digital Twin~\cite{pan2023aria}, Taskonomy~\cite{zamir2018taskonomy}, Tartanair~\cite{tartanair2020iros}, Mapfree~\cite{arnold2022mapfree}, SceneNet RGB-D~\cite{mccormac2016scenenet}, MatrixCity~\cite{li2023matrixcity}. They span indoor/outdoor, synthetic/real-world, and different scene scales.
For sequential datasets, we directly sample a whole consecutive image sequence as input. For unordered datasets, we randomly sample images observing the same scene and shuffle them as input. This approach ensures that the model can effectively learn from both structured and unstructured data inputs.

\noindent\textbf{Training objectives. }
Following VGGT~\cite{wang2025vggt}, we train our model using the multi-task loss:
\begin{align}
    \mathcal{L} = \lambda \mathcal{L}_{cam} + \mathcal{L}_{dpt} + \mathcal{L}_{xyz}
\end{align}
where $\mathcal{L}_{cam}$ denotes the L1 loss supervising the camera head, while $\mathcal{L}_{dpt}$ and $\mathcal{L}_{xyz}$ combine confidence-weighted terms with gradient-based regularisation to supervise the depth and point-cloud heads, respectively.

\noindent\textbf{Implementation details. }
We jointly train the GCM modules and the VGGT backbone end-to-end.
We use AdamW optimizer with a peak learning rate of $1\times10^{-4}$ for GCM and $1\times10^{-5}$ for the backbone. The learning rates follow a cosine decay with a 2k-iteration linear warm-up, and we apply gradient clipping with a max norm of 1.0.
Training runs for 60k iterations on 32 A800 GPUs and completes in about 3 days.
To improve length generalization, at each iteration, we randomly partition the 32 GPUs into different groups, each group processes different sequences and performs global context synchronization (GCS) only within the group, resulting in variable effective sequence lengths spanning from 1 to 32 chunks during training.

\subsection{Inference}
\label{subsec:inference}

During inference, as described in Section~\ref{subsec:architecture}, we first divide the input image set into overlapping chunks and assign them to multiple GPUs for parallel processing. Each GPU processes its local chunk through our model individually, while our Global Context Synchronization (GCS) mechanism (Section~\ref{subsec:ttt_as_memory}) enables communication of sequence-wide context across devices.

After obtaining the 3D predictions for each chunk, we follow VGGT-Long~\cite{deng2025vggt} to align and fuse results across chunks.
Specifically, we exploit the overlapping regions between adjacent chunks to compute similarity transformations for point-cloud alignment, then merge all chunks into the final kilometre-scale 3D reconstruction.
For trajectories with revisits, we additionally use retrieval-based loop candidate discovery followed by pose-graph refinement to reduce global drift.
Note that our method can also run on a single GPU by processing chunks sequentially, albeit with increased inference time.

\section{Experiment}
\label{sec:experiment}

\begin{figure*}[htbp]
    \centering
    \includegraphics[width=1.0\linewidth]{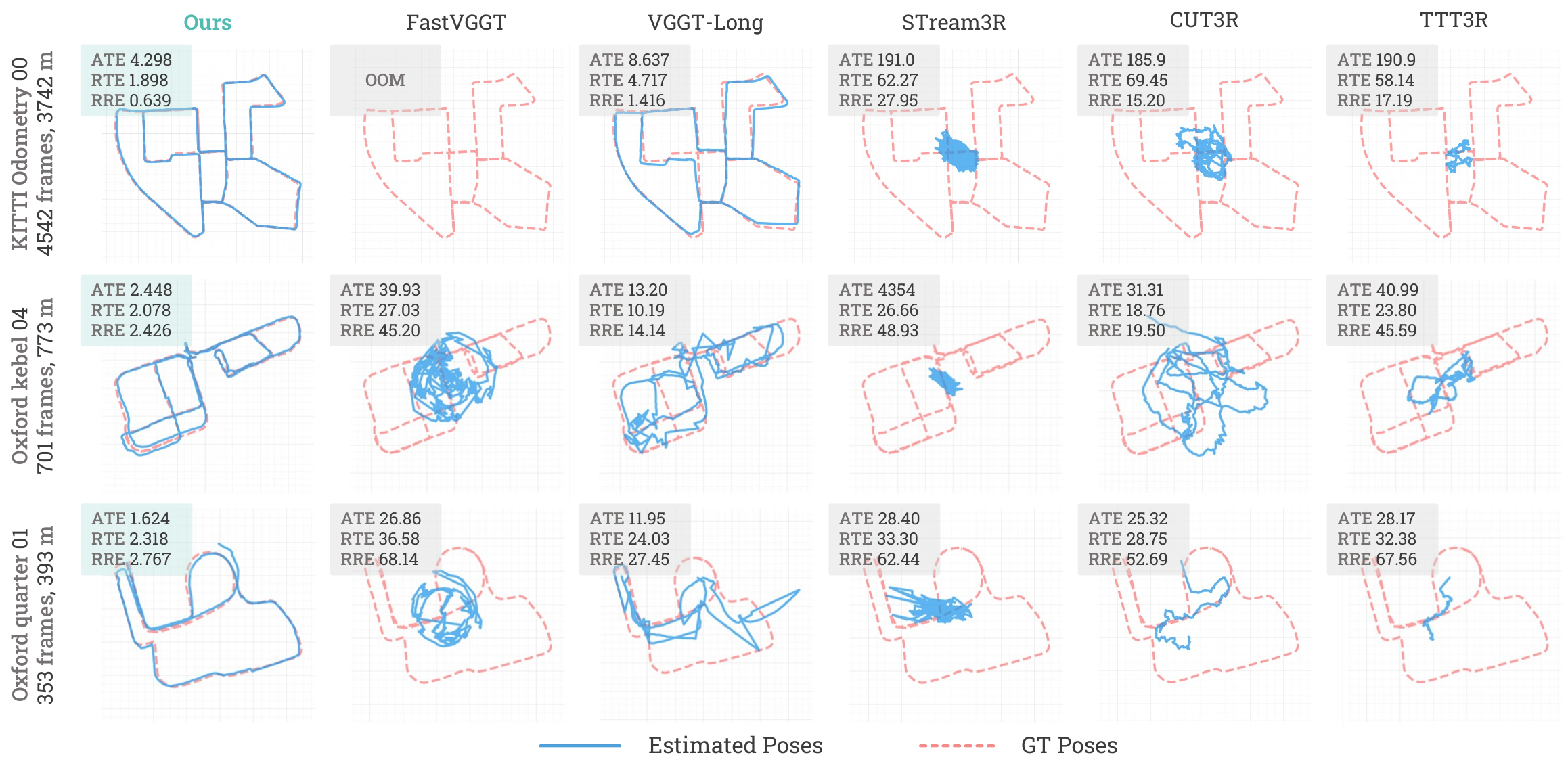}
    \vspace{-20pt}
    \caption{\textbf{Camera trajectory comparison} on KITTI Odometry~\cite{Geiger2012CVPR} and Oxford Spires~\cite{tao2025spires}. \sysname{} preserves global structure with lower drift, whereas baselines often lose tracking or diverge on long sequences.}
    \label{fig:camera_trajectory_comparison}
    \vspace{-10pt}
\end{figure*}

We evaluate \sysname{} on multiple benchmarks, including long sequence pose accuracy and 3D reconstruction accuracy. In addition, we conduct several ablation studies to analyze the impact of our key design choices.

\subsection{Pose Accuracy}
\label{subsec:pose_accuracy}

\noindent\textbf{Datasets and metrics. }
We evaluate pose accuracy on three representative datasets: Virtual KITTI (v2.0.3)~\cite{cabon2020virtual}, KITTI Odometry~\cite{Geiger2012CVPR}, and Oxford Spires~\cite{tao2025spires}.
Virtual KITTI~\cite{cabon2020virtual} is an in-domain synthetic dataset with 5 sequences spanning diverse weather and lighting conditions.
The other two are out-of-domain, real-world benchmarks:
KITTI Odometry~\cite{Geiger2012CVPR}, containing 11 sequences collected from urban driving scenarios with varying lengths, and
Oxford Spires~\cite{tao2025spires}, consisting of 6 sequences with challenging loop closures across indoor and outdoor scenes.
We report the Absolute Trajectory Error (ATE), Relative Rotation Error (RRE), and Relative Translation Error (RTE) after Sim(3) alignment with the ground truth. Details of the evaluation protocol are provided in the supplementary material.
We further report extended pose comparisons on two dense long-video benchmarks, ScanNet++~\cite{yeshwanth2023scannet++} and TUM-RGBD~\cite{sturm2012benchmark}, as well as Waymo~\cite{sun2020scalability} for outdoor driving scenes. Detailed results are provided in Table~\ref{tab:supp_pose_additional} of the supplementary material.

\begin{figure*}[t]
    \centering
    \includegraphics[width=1.0\linewidth]{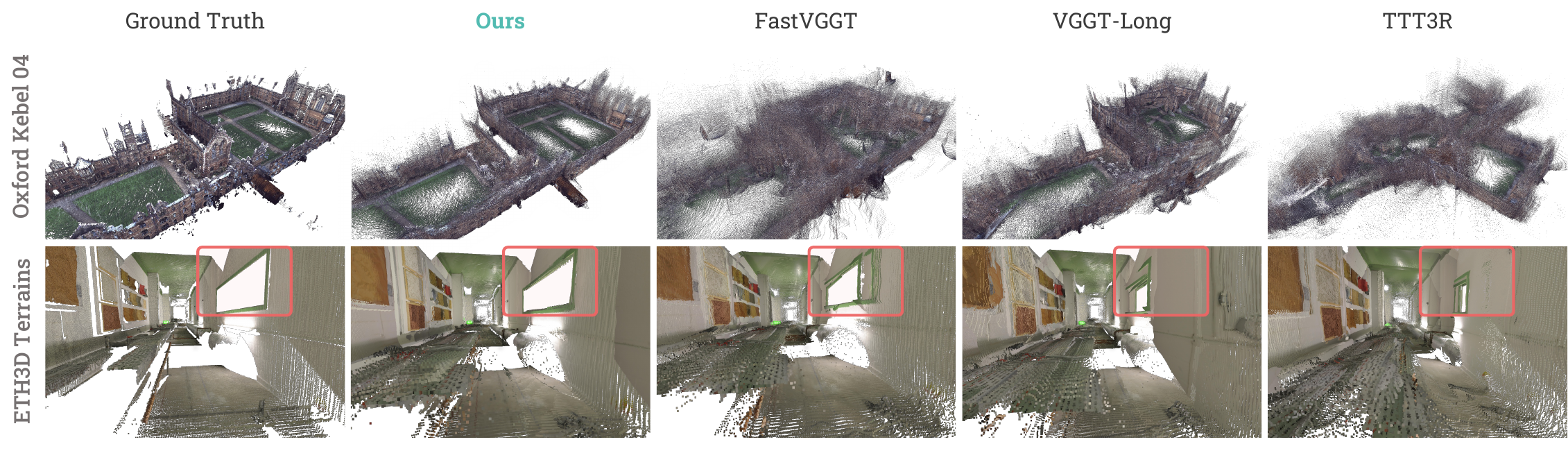}
    \vspace{-15pt}
    \caption{\textbf{Qualitative comparison of point-cloud reconstruction} on outdoor and indoor scenes. \sysname{} reconstructs large-scale outdoor scenes more reliably and preserves more consistent local geometry indoors.}
    \label{fig:comparison_point}
    \vspace{-10pt}
\end{figure*}

\begin{table}[ht!]
    \centering
    \renewcommand{\arraystretch}{0.8}
    \setlength{\tabcolsep}{2.8pt}
    \small

    \caption{\textbf{3D reconstruction evaluation} on ETH3D~\cite{Schops_2019_CVPR}, Oxford Spires~\cite{tao2025spires}, and Virtual KITTI~\cite{cabon2020virtual}. We report Chamfer Distance (CD) and F1 score. Best and second-best results are shown in \textbf{bold} and \underline{underlined}.}
    \label{tab:recon}

    \begin{tabular}{lcccccc}
        \toprule
        \multirow{2}{*}[-0.5ex]{\textbf{Method}} & \multicolumn{2}{c}{\textbf{ETH3D}}          & \multicolumn{2}{c}{\textbf{Oxford Spires}} & \multicolumn{2}{c}{\textbf{VKITTI2}}      \\
                                                 \cmidrule(lr){2-3} \cmidrule(lr){4-5} \cmidrule(lr){6-7}
                                                 & CD $\downarrow$                             & F1 $\uparrow$                              & CD $\downarrow$                              & F1 $\uparrow$                              & CD $\downarrow$                              & F1 $\uparrow$                              \\ \midrule
        MASt3R-SLAM~\cite{murai2025mast3r}       & 0.89                                        & 0.31                                       & 7.78                                         & 0.53                                       & 17.08                                       & 0.33                                       \\
        VGGT-SLAM~\cite{maggio2025vggt}          & 0.78                                        & 0.72                                       & 10.16                                        & 0.22                                       & 9.74                                        & 0.57                                       \\
        StreamVGGT~\cite{streamVGGT}             & 1.86                                        & 0.14                                       & 12.23                                        & 0.25                                       & 20.45                                       & 0.35                                       \\
        STream3R~\cite{lan2025stream3r}          & 1.81                                        & 0.14                                       & 12.20                                        & 0.25                                       & 18.77                                       & 0.36                                       \\
        CUT3R~\cite{wang2025continuous}          & 0.41                                        & 0.60                                       & 6.93                                         & 0.45                                       & 5.67                                        & 0.39                                       \\
        TTT3R~\cite{chen2025ttt3r}               & 0.43                                        & 0.59                                       & 9.03                                         & 0.31                                       & 3.49                                        & 0.49                                       \\
        FastVGGT~\cite{shen2025fastvggt}         & 0.50                                        & 0.70                                       & \underline{2.76}                             & 0.76                                       & \underline{1.73}                            & 0.67                                       \\
        VGGT-Long~\cite{deng2025vggt}            & \underline{0.24}                            & \underline{0.84}                           & 3.41                                         & \underline{0.80}                           & 1.78                                        & \underline{0.70}                           \\ \midrule
        Ours                                     & \textbf{0.11}                               & \textbf{0.91}                              & \textbf{0.96}                                & \textbf{0.96}                              & \textbf{0.40}                               & \textbf{0.91}                              \\ \bottomrule
    \end{tabular}

    \vspace{-15pt}
\end{table}

\noindent\textbf{Baseline comparisons. }
We compare our \sysname{} against extensive baselines, including VGGT-Long~\cite{deng2025vggt}, FastVGGT~\cite{shen2025fastvggt}, foundation models with memory mechanisms CUT3R~\cite{wang2025continuous}, STream3R~\cite{lan2025stream3r}, StreamVGGT~\cite{streamVGGT}, TTT3R~\cite{chen2025ttt3r}, and recent learning-based SLAM methods MASt3R-SLAM~\cite{murai2025mast3r} and VGGT-SLAM~\cite{maggio2025vggt}. We also include SfM baselines COLMAP~\cite{schonberger2016structure}, MASt3R-SfM~\cite{leroy2024grounding} and classical SLAM baselines DROID-SLAM~\cite{teed2021droid}, DPVO++~\cite{lipson2024deep}, which assume known camera intrinsics.
As shown in Table~\ref{tab:pose} and Figure~\ref{fig:camera_trajectory_comparison}, \sysname{} consistently outperforms feed-forward and streaming baselines across the reported metrics.
On long, challenging sequences, several baselines suffer from tracking failures (e.g., MASt3R-SLAM~\cite{murai2025mast3r}, VGGT-SLAM~\cite{maggio2025vggt}) or out-of-memory errors (e.g., FastVGGT~\cite{shen2025fastvggt}).
Notably, while TTT3R~\cite{chen2025ttt3r} improves over CUT3R~\cite{wang2025continuous} by introducing online learning for better memory updates, it still struggles on long sequences due to limited memory capacity.
Even the most competitive baseline VGGT-Long~\cite{deng2025vggt}, while strong on KITTI Odometry, lags behind \sysname{} and degrades notably on other datasets.
Classical SfM can remain competitive when feature matching and global optimization are well conditioned, as seen from COLMAP on Oxford Spires. However, it degrades on the longer, larger-scale video benchmarks considered here and is extremely slow.
These results validate the effectiveness of our global context and aggregation mechanism in capturing long-term dependencies, yielding substantial gains in pose estimation accuracy.

\noindent\textbf{Resource comparison. }
The last three columns of Table~\ref{tab:pose} report peak GPU memory, total inference time, and throughput on KITTI sequences 03, 04, and 10 (avg. 758 frames). All methods run on a single RTX 4090 except FastVGGT, which requires an A800.
Compared with FastVGGT, \sysname{} remains practical on a single GPU with moderate memory consumption while avoiding the substantial memory growth of long-context models.
Although lightweight online systems such as DPVO++ and CUT3R achieve higher throughput, our method provides substantially stronger accuracy on long sequences, while COLMAP is over $20\times$ slower than \sysname{}.
Runtime scaling with sequence length is further analyzed in the supplementary material, where runtime grows smoothly while Relative Pose Error (RPE) remains stable.

\subsection{Geometry Accuracy}
\label{subsec:geometry_accuracy}

\noindent\textbf{Datasets and metrics. }
We evaluate 3D reconstruction on ETH3D~\cite{Schops_2019_CVPR}, Virtual KITTI~\cite{cabon2020virtual}, and Oxford Spires~\cite{tao2025spires}, with 11, 50, and 6 scenes, respectively. These datasets cover diverse indoor and outdoor environments with varying scales and complexities.
We report Chamfer distance and F1 score on point clouds reconstructed from the predicted poses and depth maps, after aligning them to the ground truth using the Umeyama algorithm~\cite{umeyama2002least}.

\noindent\textbf{Baseline comparisons. }
Under the same setting as Sec.~\ref{subsec:pose_accuracy}, \sysname{} achieves strong geometric accuracy in Table~\ref{tab:recon}.
As in pose evaluation, methods that encounter tracking failures, OOM errors, or large pose deviations typically cannot produce valid reconstructions.
Moreover, performance on ETH3D~\cite{Schops_2019_CVPR} demonstrates good transfer to shorter indoor sequences, indicating the robustness of \sysname{}.
Figure~\ref{fig:comparison_point} illustrates qualitative comparisons, where \sysname{} produces more accurate large-scale reconstructions and more consistent local geometry.

\subsection{Ablation Study}
\label{subsec:ablation_study}

We ablate two key design choices in \sysname{}: the state size of the lightweight sub-networks and the global context design. All ablation models are trained and evaluated on a subset of the datasets listed in Sec.~\ref{subsec:training}, as detailed in the supplementary material.

\noindent\textbf{State size of sub-networks. }
Increasing the lightweight sub-network state size from 1M to 4M improves ATE, RTE, and RRE in the left block of Table~\ref{tab:ablation}, suggesting that larger state capacity helps preserve long-range context.

\begin{table}[ht!]
    \centering
    \renewcommand{\arraystretch}{0.84}
    \setlength{\tabcolsep}{2.7pt}
    \small

    \caption{\textbf{Ablation studies.} Left: varying GCM state size. Right: ablating global context on a complementary long-sequence setting. The two blocks are not directly comparable. Best and second-best results are shown in \textbf{bold} and \underline{underlined}.}
    \label{tab:ablation}

    \resizebox{0.97\columnwidth}{!}{%
    \begin{tabular}{l c c c l c c c}
        \toprule
        \multicolumn{4}{c}{\textbf{State Size}} & \multicolumn{4}{c}{\textbf{Global Context}} \\
        \cmidrule(lr){1-4} \cmidrule(lr){5-8}
        \textbf{Variants}         & RRE $\downarrow$      & RTE $\downarrow$      & ATE $\downarrow$      & \textbf{Variants}    & RRE $\downarrow$      & RTE $\downarrow$      & ATE $\downarrow$      \\
        \midrule
        1M size                   & 1.01                  & 1.01                  & 0.99                  & w/o GCM              & 1.30                  & 7.03                  & 19.00                 \\
        2M size                   & \underline{0.95}      & \underline{0.91}      & \underline{0.93}      & w/o GCS              & \underline{1.28}      & \underline{7.01}      & \underline{15.80}     \\
        4M size                   & \textbf{0.87}         & \textbf{0.84}         & \textbf{0.85}         & Full model           & \textbf{1.17}         & \textbf{5.99}         & \textbf{13.70}        \\
        \bottomrule
    \end{tabular}
    }
\end{table}

\noindent\textbf{Global context mechanism. }
We also ablate `w/o GCS', which removes cross-chunk context synchronization, and `w/o GCM', which removes the global context memory. In the right block of Table~\ref{tab:ablation}, both variants worsen ATE relative to the full model, with the larger drop for `w/o GCM' showing that GCM carries primary long-range context while GCS helps propagate it across chunks.

\section{Conclusion}
\label{sec:conclusion}

We present \sysname{}, a scalable framework for 3D reconstruction from long RGB sequences. It combines neural global context with online-adapted lightweight sub-networks and context aggregation to preserve long-range dependencies efficiently. Extensive experiments demonstrate \sysname{}'s state-of-the-art pose estimation and 3D geometry accuracy.

\noindent\textbf{Acknowledgment.} This work was partially supported by National Key R\&D Program of China (No.~2024YFB2809105), NSFC (No.~U24B20154), and Information Technology Center and State Key Lab of CAD\&CG, Zhejiang University. We thank Tianyuan Zhang for helpful discussions on LaCT and Dongli Tan for valuable discussions.

{
    \small
    \bibliographystyle{ieeenat_fullname}
    \bibliography{main}
}

\clearpage
\setcounter{page}{1}
\maketitlesupplementary
\appendix

\section{Model Details}
\label{supsec:model_details}

We provide the detailed model architecture in this section.

\noindent\textbf{Overall architecture.}
As stated in Section~\ref{subsec:architecture}, we build our model up as a large transformer, as VGGT~\cite{wang2025vggt}.
We then attach a Global Context Memory (GCM) module after 4 specific global attention layers, namely 4th, 11th, 17th, and 24th, whose outputs are used as input features of the two DPT~\cite{ranftl2021vision} decoders to predict the depth maps and point clouds.
The total number parameters of the newly added GCM module is 75.55M, namely 0.076B.

\noindent\textbf{Global Context Memory module.}
The GCM module consists of three components: a query-key-value projection layer, three compact MLP networks $W_1, W_2, W_3$ serving as the Adaptive Memory Units (AMUs), and an output projection layer.
The forward pass of the GCM module is performed as follows:
\begin{align}
    f_W(x) = W_2 \big( \mathrm{SiLU}(W_1x) \circ (W_3x) \big),
\end{align}
where $\circ$ denotes the element-wise product, after we update the AMUs $W_1, W_2, W_3$ in the inner loop using $K, V$ as detailed in the Section~\ref{subsec:ttt_as_memory}, we can use the updated AMUs to compute the GCM output $f_W(Q)$.
The query-key-value projection layer is the standard linear projection layer, which projects the upstream feature $x\in\mathbb{R}^{M\times d}$ into multi-head query $Q\in\mathbb{R}^{M \times nh \times hd}$, key $K\in\mathbb{R}^{M \times nh \times hd}$, and value $V\in\mathbb{R}^{M \times nh \times hd}$, where $nh$ is the number of heads and $hd$ is the dimension of each head, where $nh \times hd = d$.
Define the hidden dimension of the AMUs as $hd \times k$, with $k$ being a scaling factor, then $W_1, W_3 \in \mathbb{R}^{hd \times hd \times k}$ and $W_2 \in \mathbb{R}^{hd \times k \times hd}$.
The total state size of the GCM module is calculated as:
\begin{align}
    \text{state size} = nh \times hd \times hd \times k = \frac{d^2}{nh} \times k.
\end{align}
Specifically, we set the number of heads $nh$ to 1 to maximize the state size for larger memory capacity, and set the scaling factor $k$ to 4 to balance the memory capacity and computational efficiency.

\section{Evaluation Details}
\label{supsec:evaluation_details}

\subsection{Dataset Details}
\label{supsubsec:dataset_details}

Our benchmarks are built on four datasets: Virtual KITTI~\cite{cabon2020virtual}, KITTI Odometry~\cite{Geiger2012CVPR}, Oxford Spires~\cite{tao2025spires}, and ETH3D~\cite{Schops_2019_CVPR}. These datasets feature long, large-scale sequences with diverse weather and lighting conditions, urban driving scenarios, and indoor and outdoor scenes, respectively. We present more details about the datasets in the following.

\noindent\textbf{Virtual KITTI}
\cite{cabon2020virtual} is a synthetic dataset comprising 50 outdoor street-scene sequences spanning diverse weather and lighting conditions (e.g., fog, morning, overcast, overcast, rain and sunset).
Sequence lengths range from 223–837 frames, with path lengths spanning 52–711 meters.

\noindent\textbf{KITTI Odometry}
\cite{Geiger2012CVPR} is a real-world benchmark of 11 sequences collected from urban driving scenarios with varied lengths and street layouts.
Sequence lengths range from 271–4,661 frames, covering 0.39–5.07 km of travel, and pose challenging long-sequence tracking conditions.

\noindent\textbf{Oxford Spires}
\cite{tao2025spires} is a real-world dataset with 6 sequences, \texttt{2024-03-12-keble-college-02}, \texttt{2024-03-12-keble-college-03}, \texttt{2024-03-12-keble-college-04}, \texttt{2024-03-12-keble-college-05}, \texttt{2024-03-13-observatory-quarter-01}, \texttt{2024-03-13-observatory-quarter-02}, featuring challenging loop closures and extreme view sparsity across indoor and outdoor scenes.
Sequence lengths range from 351–787 frames, covering 280–773 meters.
To ensure reliable supervision and fair evaluation, we filtered out views with large LiDAR–RGB timestamp discrepancies and removed scenes that consequently contained fewer than 50 frames despite spanning several hundred meters.

\noindent\textbf{ETH3D}
\cite{Schops_2019_CVPR} provides high-resolution indoor and outdoor images with ground-truth depth from laser sensors. We select 11 scenes: \texttt{courtyard}, \texttt{electro}, \texttt{kicker}, \texttt{pipes}, \texttt{relief}, \texttt{delivery area}, \texttt{facade}, \texttt{office}, \texttt{playground}, \texttt{relief 2}, \texttt{terrains}, for the benchmark. The number of frames in each scene ranges from 14 to 76.

\subsection{Evaluation Details}
\label{supsubsec:evaluation_details}

We provide the detailed evaluation in this section.

\noindent\textbf{Pose metrics. }
For pose accuracy evaluation, we follow the protocol introduced in \cite{lan2025stream3r,deng2025vggt} , and report results using the Absolute Trajectory Error (ATE), Relative Rotation Error (RRE in $^\circ$/100m), and Relative Translation Error (RTE in m/100m), providing a comprehensive assessment of both translation and rotation accuracy.
All metrics are calculated after Sim(3) alignment of predicted pose trajectories with the ground truth.

\begin{figure*}[ht!]
    \centering
    \includegraphics[width=1.0\linewidth]{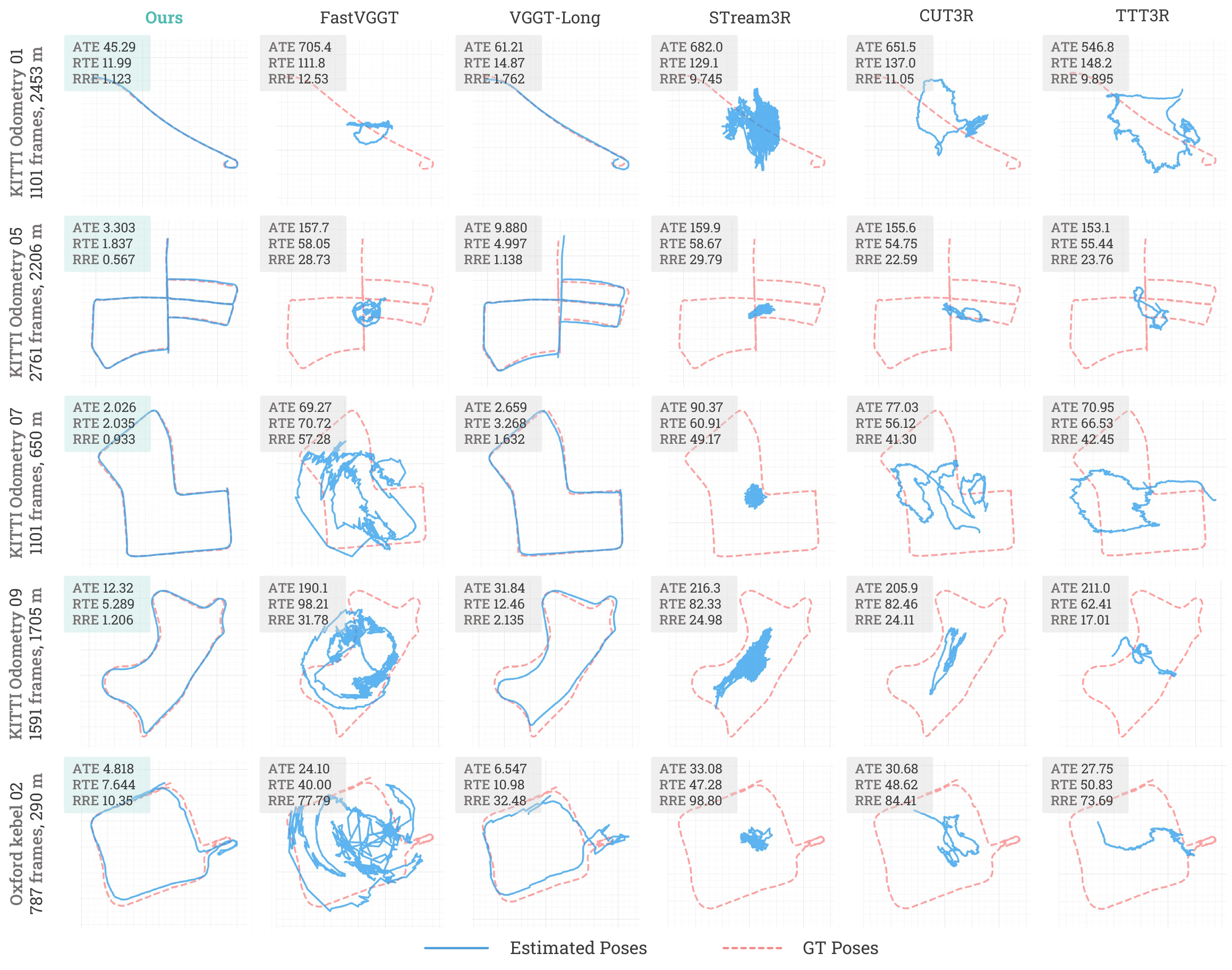}
    \caption{\textbf{Camera trajectory comparison.} \sysname{} preserves global structure with substantially lower drift, whereas baselines frequently lose tracking or diverge, demonstrating our capability of reconstructing large-scale scenarios with high accuracy.}
    \label{fig:supp_camera_trajectory_comparison}
\end{figure*}

\noindent\textbf{Reconstruction metrics. }
We evaluate 3D reconstruction with Chamfer Distance (CD) and F1-score.
Let $\mathcal{G}$ be the ground-truth point cloud and $\mathcal{P}$ the predicted point cloud after Sim(3) alignment with the ground-truth using the Umeyama algorithm~\cite{umeyama2002least}.
Denote by $\operatorname{dist}(A \rightarrow B)$ the average nearest-neighbour distance from each point in $\mathcal{A}$ to $\mathcal{B}$.
We define accuracy as $\operatorname{dist}(\mathcal{P} \rightarrow \mathcal{G})$ and completeness as $\operatorname{dist}(\mathcal{G} \rightarrow \mathcal{P})$, then the Chamfer Distance (CD) is defined as the average of accuracy and completeness.
Given a distance threshold $d$, we define the precision and recall as:
\begin{align}
    \text{precision} &= \frac{1}{|\mathcal{P}|} \sum_i [\operatorname{dist}(\mathcal{P}_i \rightarrow \mathcal{G}) < d], \\
    \text{recall} &= \frac{1}{|\mathcal{G}|} \sum_i [\operatorname{dist}(\mathcal{G}_i \rightarrow \mathcal{P}) < d],
\end{align}
where $[\cdot]$ denotes the Iverson bracket~\cite{knapitsch2017tanks}. Then, the F1-score is computed as:
\begin{align}
    \mathrm{F1} = \frac{2 \times \text{precision} \times \text{recall}} {\text{precision} + \text{recall}}.
\end{align}

\begin{figure*}[ht!]
    \centering
    \includegraphics[width=1.0\linewidth]{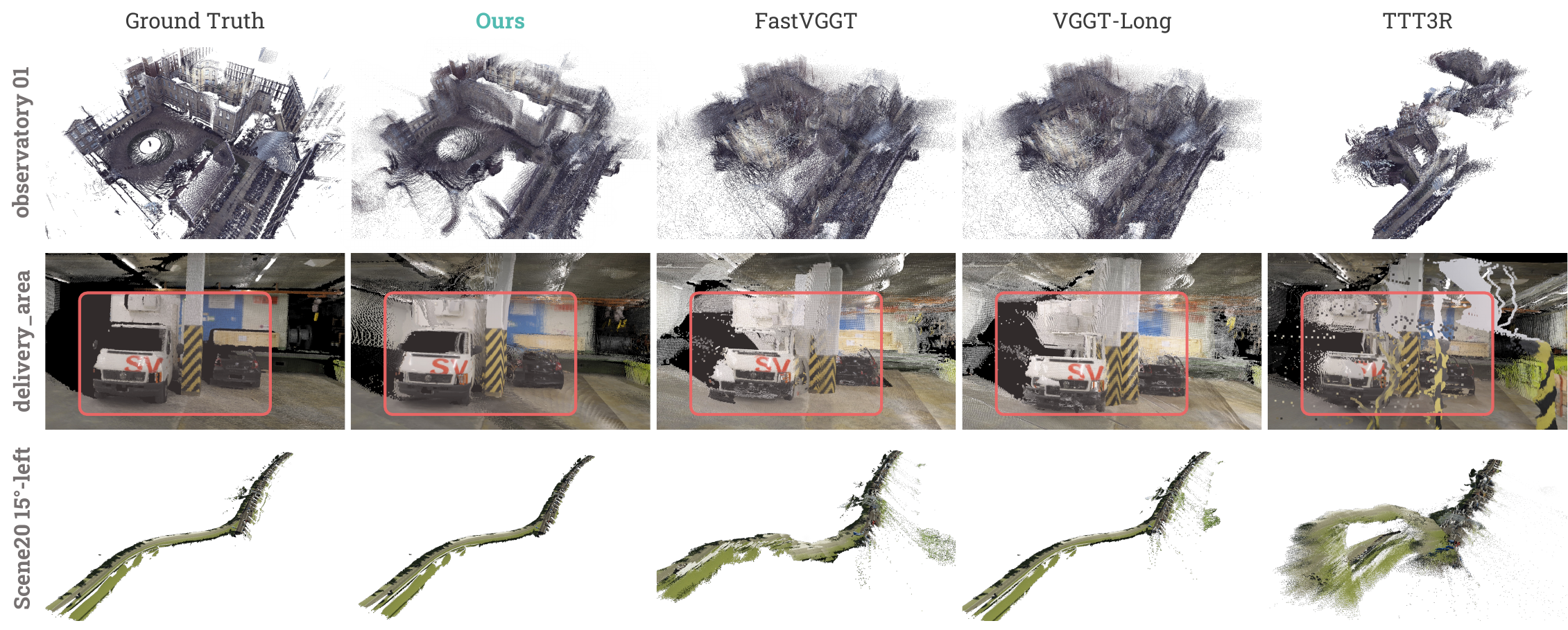}
    \caption{\textbf{Point-cloud reconstruction comparison.} \sysname{} produces more accurate large-scale reconstructions for large-scale outdoor environments where baselines often fail, and achieves higher local geometric accuracy and consistency in indoor scenes.}
    \label{fig:supp_point_cloud_reconstruction_comparison}
\end{figure*}

\noindent\textbf{Evaluation details. }
All evaluations use the full set of frames per sequence.
We set chunk size to 60 and overlap to 30 for both \sysname{} and VGGT-Long~\cite{deng2025vggt} across all datasets, and we follow official evaluation protocols for the remaining baselines~\cite{shen2025fastvggt,wang2025continuous,lan2025stream3r,streamVGGT,chen2025ttt3r,murai2025mast3r,maggio2025vggt}.
Pose metrics are directly evaluated on Virtual KITTI~\cite{cabon2020virtual}, KITTI Odometry~\cite{Geiger2012CVPR}, and Oxford Spires~\cite{tao2025spires} datasets with no extra hyperparameters.
Reconstruction metrics are evaluated on ETH3D~\cite{Schops_2019_CVPR}, Virtual KITTI~\cite{cabon2020virtual}, and Oxford Spires~\cite{tao2025spires} datasets, using dataset-specific distance thresholds (ETH3D: 0.25, Virtual KITTI: 1.0, Oxford Spires: 4.0) to reflect differences in scale and sparsity.
For baselines that fail to produce valid camera trajectories or reconstructions on a scene, we assign the worst valid score among the compared methods on that scene when computing dataset averages in Sections~\ref{subsec:pose_accuracy} and~\ref{subsec:geometry_accuracy}.

\noindent\textbf{Ablation details. }
To simplify training while preserving the generality of our conclusions, we train all ablation models on a subset of the datasets listed in Section~\ref{subsec:training}, excluding the object-centric datasets WildRGB~\cite{xia2024rgbd}, Co3Dv2~\cite{reizenstein2021common}, and Aria Digital Twin~\cite{pan2023aria}.
State size ablations are trained on 16 NVIDIA A800 GPUs for 60k iterations for a fair comparison.
We randomly select 7 sequences from Virtual KITTI~\cite{cabon2020virtual} (\texttt{Scene01 15-deg-left}, \texttt{Scene02 30-deg-left}, \texttt{Scene06 clone}, \texttt{Scene18 morning}, \texttt{Scene20 rain}), Oxford Spires~\cite{tao2025spires} (\texttt{2024-03-12-keble-college-04}, \texttt{2024-03-13-observatory-quarter-01}) for evaluation, covering diverse weather and lighting conditions, urban driving scenarios, and indoor and outdoor scenes.
Global context ablations are trained on 8 NVIDIA A800 GPUs for 60k iterations for a fair comparison.
We select KITTI Odometry~\cite{Geiger2012CVPR} sequences \texttt{01}, \texttt{03}, \texttt{04}, \texttt{10}, and Virtual KITTI~\cite{cabon2020virtual} sequences \texttt{Scene20} for evaluation, featuring challenging long-sequence tracking conditions.

\section{Additional Results}
\label{supsec:additional_results}

We provide additional long-sequence camera trajectory results in Figure~\ref{fig:supp_camera_trajectory_comparison}.
As demonstrated in Figure~\ref{fig:supp_camera_trajectory_comparison}, our method is capable of reconstructing extremely large-scale long sequences with small drift, whereas baselines frequently lose tracking or diverge significantly, showcasing the effectiveness of our proposed global context representation and aggregation mechanism.
We provide additional long-sequence reconstruction results in Figure~\ref{fig:supp_point_cloud_reconstruction_comparison}, the results illustrate the improvement of our method over baselines on both large-scale accurate reconstruction and local geometric consistency.

\subsection{Additional Benchmark Comparisons}
\label{supsubsec:additional_benchmark_comparisons}

We further evaluate pose accuracy on three additional benchmarks in Table~\ref{tab:supp_pose_additional}. For ScanNet++~\cite{yeshwanth2023scannet++}, we use five sequences: \texttt{419cbe7c11}, \texttt{98b4ec142f}, \texttt{bb87c292ad}, \texttt{c24f94007b}, and \texttt{ebc200e928}. For TUM-RGBD~\cite{sturm2012benchmark}, we evaluate all scenes. For Waymo~\cite{sun2020scalability}, we follow VGGT-Long~\cite{deng2025vggt} and use the same nine test scenes. Compared with the main paper benchmarks, these datasets are denser video regimes with stronger short-range overlap, so they are useful for checking whether our gains persist when recent streaming and video-based baselines are relatively better matched to the evaluation setting.
We follow the same evaluation protocol as in the main paper and report ATE after Sim(3) alignment with the ground truth.
We set the chunk size and overlap to 120 and 60, respectively, for both \sysname{} and VGGT-Long~\cite{deng2025vggt} on ScanNet++~\cite{yeshwanth2023scannet++} and TUM-RGBD~\cite{sturm2012benchmark}, and to 60 and 30, respectively, on Waymo~\cite{sun2020scalability}.
As shown in Table~\ref{tab:supp_pose_additional}, \sysname{} achieves the best ATE on ScanNet++ (0.08) and TUM-RGBD (0.07), with clear margins over strong video-based baselines such as STream3R and TTT3R. This shows that the proposed global context mechanism is not only helpful for the large-scale sparse settings emphasized in the main paper, but also remains effective on denser long-video benchmarks. On Waymo, \sysname{} remains competitive on long driving sequences, indicating good transfer across different video regimes.
\begin{table}[ht]
    \centering
    \small
    \setlength{\tabcolsep}{2.2pt}
    \setlength{\aboverulesep}{0.35ex}
    \setlength{\belowrulesep}{0.35ex}
    \renewcommand{\arraystretch}{1.02}
    \caption{\textbf{Additional pose benchmark comparisons.} We report ATE (m, lower is better) on three supplementary benchmarks. The best results are in \textbf{bold}, and the second best are \underline{underlined}.}
    \label{tab:supp_pose_additional}
    \resizebox{0.97\columnwidth}{!}{%
    \begin{tabular}{lccc}
        \toprule
        \multirow{2}{*}[-0.5ex]{\textbf{Method}} &
        {\small \textbf{ScanNet++}} &
        {\small \textbf{TUM-RGBD}} &
        {\small \textbf{Waymo}} \\
        & \scriptsize avg. 924 fr. & \scriptsize avg. 926 fr. & \scriptsize avg. 198 fr. \\
        \midrule
        MASt3R-SLAM~\cite{murai2025mast3r}         & 0.47                              & \underline{0.08}                  & 7.63                              \\
        VGGT-SLAM~\cite{maggio2025vggt}            & 0.29                              & 0.12                              & 7.43                              \\
        StreamVGGT~\cite{streamVGGT}               & 1.70                              & 0.63                              & 45.10                             \\
        STream3R~\cite{lan2025stream3r}            & 1.75                              & 0.63                              & 42.20                             \\
        CUT3R~\cite{wang2025continuous}            & 1.27                              & 0.54                              & 9.40                              \\
        TTT3R~\cite{chen2025ttt3r}                 & 0.55                              & 0.31                              & 3.49                              \\
        FastVGGT~\cite{shen2025fastvggt}           & 1.56                              & 0.42                              & \textbf{1.28}                     \\
        VGGT-Long~\cite{deng2025vggt}              & \underline{0.13}                  & \underline{0.08}                  & 1.78                              \\
        COLMAP~\cite{schonberger2016structure}     & GT                                & 0.19                              & 25.63                             \\
        MASt3R-SfM~\cite{leroy2024grounding}       & 1.50                              & 0.39                              & 3.95                              \\
        DROID-SLAM~\cite{teed2021droid}            & 0.97                              & 0.11                              & 6.67                              \\
        DPVO++~\cite{lipson2024deep}               & 0.91                              & 0.10                              & \underline{1.35}                  \\
        \midrule
        \textbf{Ours}                              & \textbf{0.08}                     & \textbf{0.07}                     & 1.52                              \\
        \bottomrule
    \end{tabular}%
    }
\end{table}

\subsection{Runtime Scaling with Sequence Length}
\label{supsubsec:runtime_scaling}

We further analyze runtime scaling with sequence length in Table~\ref{tab:supp_runtime_scaling}. As the sequence length increases from 150 to 990 frames, the total runtime grows approximately linearly, while throughput remains stable at around 2.6--2.9 FPS. Meanwhile, the relative pose error remains within 0.07--0.08 m, indicating that \sysname{} maintains stable pose accuracy as the sequence length increases.
\begin{table}[ht]
    \centering
    \small
    \setlength{\tabcolsep}{5.2pt}
    \setlength{\aboverulesep}{0.35ex}
    \setlength{\belowrulesep}{0.35ex}
    \renewcommand{\arraystretch}{1.02}
    \caption{\textbf{Runtime scaling with sequence length.} Using the same single-GPU evaluation setting as the main-paper resource comparison, we report relative pose error (RPE, m), total inference time, and FPS as the input length increases. Runtime grows smoothly with sequence length while RPE remains stable.}
    \label{tab:supp_runtime_scaling}
    \begin{tabular}{lcccc}
        \toprule
        \textbf{Frames}                & 150       & 270       & 510       & 990       \\
        \midrule
        \textbf{RPE (m)} $\downarrow$  & 0.08      & 0.08      & 0.07      & 0.08      \\
        \textbf{Time (s)} $\downarrow$ & 51.19     & 98.81     & 195.24    & 382.80    \\
        \textbf{FPS} $\uparrow$        & 2.93      & 2.73      & 2.61      & 2.59      \\
        \bottomrule
    \end{tabular}
\end{table}

\subsection{Failure Cases}
\label{supsubsec:failure_cases}

\begin{figure}[H]
    \centering
    \includegraphics[width=1.0\columnwidth]{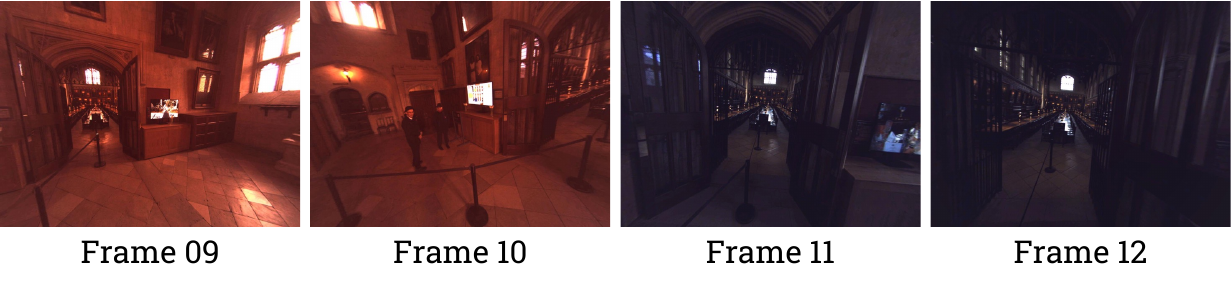}
    \caption{\textbf{Failure case under abrupt illumination changes.} Large appearance shifts within a sequence weaken cross-chunk correspondences and can lead to inaccurate global alignment.}
    \label{fig:supp_failure_case_illumination}
\end{figure}

We further summarize representative failure modes of \sysname{}.
The first arises from severe appearance inconsistency within a sequence (e.g., abrupt illumination or color shifts), as illustrated in Figure~\ref{fig:supp_failure_case_illumination}. In such cases, the appearance gap across chunks weakens the reliability of cross-chunk correspondences.
The second occurs under extreme view sparsity, for example when only tens of images cover scenes spanning hundreds of meters or even kilometers. In such extreme cases, even local predictions can fail due to the lack of sufficient geometric constraints.

% {
%     \small
%     \bibliographystyle{ieeenat_fullname}
%     \bibliography{supplementary}
% }

\end{document}